%% file: main.tex
\documentclass{article}

\usepackage[numbers]{natbib}
\usepackage[preprint]{neurips_2023}
\usepackage[dvipsnames]{xcolor}         
\usepackage[utf8]{inputenc} 
\usepackage[T1]{fontenc}    
\usepackage[colorlinks=true,linkcolor=linkColor,citecolor=linkColor,filecolor=linkColor,urlcolor=linkColor]{hyperref}       
\usepackage{url}            
\usepackage{booktabs}       
\usepackage{amsfonts}       
\usepackage{nicefrac}       
\usepackage{microtype}      
\usepackage{graphicx}
\usepackage{algorithm}
\usepackage{algorithmic}
\usepackage{ragged2e}
\usepackage{amsmath}
\usepackage{array}          
\usepackage{tabularx}
\usepackage{colortbl}
\usepackage{float}
\usepackage{comment}

\usepackage{diagbox}

\usepackage{color}
\usepackage{colortbl}
\usepackage[most]{tcolorbox}
\usepackage{wrapfig}
\usepackage{soul}
\usepackage{lipsum}
\usepackage{xcolor}

\usepackage{tablefootnote}

\definecolor{linkColor}{rgb}{0.18,0.39,0.62}

\tcbset {
  base/.style={
    arc=0mm, 
    bottomtitle=-0.25mm,
    boxrule=0.5mm,
    colbacktitle=black!10!white, 
    coltitle=black, 
    fonttitle=\bfseries, 
    left=2.5mm,
    right=3.5mm,
    title={#1},
    toptitle=0.25mm,
  }
}

\definecolor{brandblue}{rgb}{0.34, 0.7, 1}
\definecolor{brandgreen}{rgb}{0.50, 0.71, 0.38}
\newtcolorbox{mybox}[1]{
  colframe=brandblue, 
  base={#1}
}
\sethlcolor{brandgreen} 

\input{settings.tex}
\input{math_commands.tex}

\title{Enhancing Language Model Rationality with Bi-Directional Deliberation Reasoning}

\author{
  Yadong Zhang \textsuperscript{1}\thanks{Work was done when interning at Microsoft Research Asia.}  ~~ Shaoguang Mao\textsuperscript{2} ~~ Wenshan Wu\textsuperscript{2} \\
  \textbf{Yan Xia}\textsuperscript{2} 
  ~~ \textbf{Tao Ge}\textsuperscript{2} 
  ~~ \textbf{Man Lan}\textsuperscript{1} ~~ \textbf{Furu Wei}\textsuperscript{2}\\
  \textsuperscript{1}\space\texttt{East China Normal University} \\
  \textsuperscript{2}\space\texttt{Microsoft Research} \\
  \href{https://aka.ms/GeneralAI}{https://aka.ms/GeneralAI}
}

\begin{document}

\maketitle
\vspace{-0.7cm}
\begin{abstract}This paper introduces \textbf{BI}-\textbf{D}irectional \textbf{DE}liberation \textbf{R}easoning (BIDDER), a novel reasoning approach to enhance the decision rationality of language models. Traditional reasoning methods typically rely on historical information and employ uni-directional (left-to-right) reasoning strategy. This lack of bi-directional deliberation reasoning results in limited awareness of potential future outcomes and insufficient integration of historical context, leading to suboptimal decisions. BIDDER addresses this gap by incorporating principles of rational decision-making, specifically managing uncertainty and predicting expected utility. Our approach involves three key processes: Inferring hidden states to represent uncertain information in the decision-making process from historical data; Using these hidden states to predict future potential states and potential outcomes; Integrating  historical information (past contexts) and long-term outcomes (future contexts) to inform reasoning. By leveraging bi-directional reasoning, BIDDER ensures thorough exploration of both past and future contexts, leading to more informed and rational decisions. We tested BIDDER's effectiveness in two well-defined scenarios: Poker (Limit Texas Hold'em) and Negotiation. Our experiments demonstrate that BIDDER significantly improves the decision-making capabilities of LLMs and LLM agents.
\end{abstract}

\section{Introduction}
The rapid advancement of large language models (LLMs) has significantly enhanced natural language understanding and generation \cite{naveed2023comprehensive}. However, as the application scope of LLMs expands, they often struggle to make rational and optimal decisions in real-world scenarios\cite{chen2023asking, duan2024gtbench, liu2024position}, particularly in complex environments that require careful consideration of both historical data and potential future outcomes.

Existing methods predominantly enhance reasoning performance by navigating through the problem space and selecting optimal nodes (states) based on the progress toward solving the problem \cite{wei2022chain, yao2023tree}. These unidirectional (left-to-right) reasoning approaches limit LLMs' ability to effectively leverage future information, resulting in suboptimal decisions due to insufficient integration of historical context with potential future outcomes \cite{havrilla2024teaching}. To address these shortcomings, we propose \textbf{BI}-\textbf{D}irectional \textbf{DE}liberation \textbf{R}easoning (BIDDER), a novel approach designed to enhance decision rationality. BIDDER leverages the strengths of classical bi-directional modeling, as exemplified by Bidirectional Encoder Representations from Transformers (BERT \cite{devlin2018bert}), and integrates these with decision theory principles to overcome the limitations of traditional reasoning methods in LLMs.

The key innovation of BIDDER lies in its bi-directional reasoning capability, which ensures a thorough exploration of both past and future contexts. As illustrated in Figure \ref{fig
}, unidirectional reasoning utilizes historical context to make left-to-right decisions. In contrast, bi-directional reasoning explores potential future states and predicts the expected utility for future moves, using both historical contexts and future exploration contexts for more informative reasoning.

\begin{wrapfigure}{r}{0.52\textwidth}
\vspace{-5mm}
\hspace{-4mm}
\includegraphics[width=1.1\linewidth]{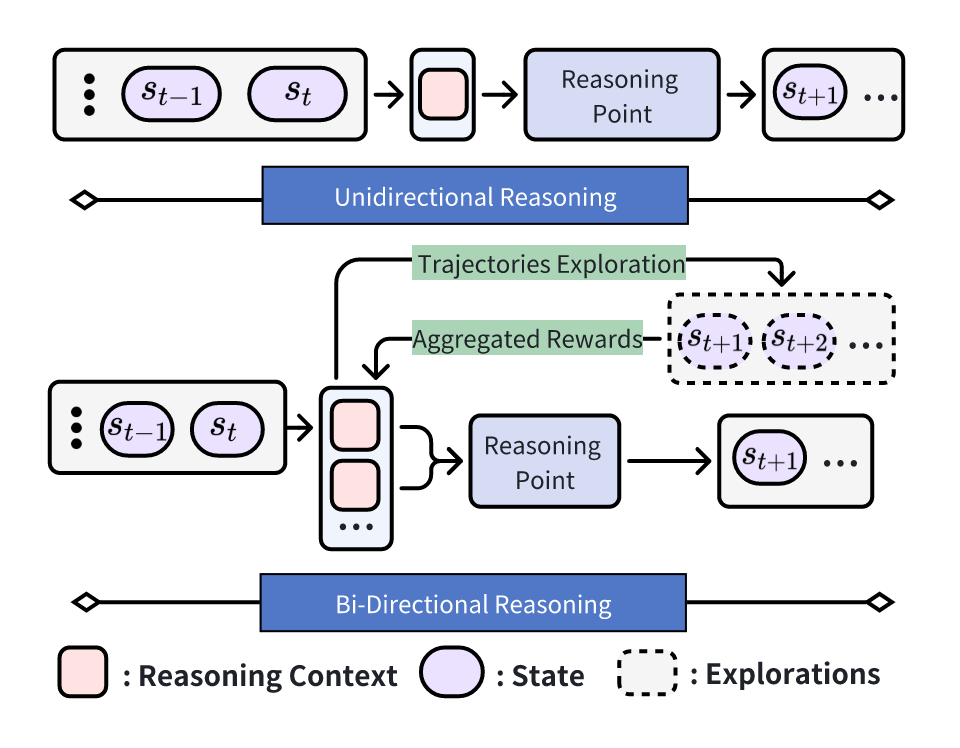}
\vspace{-0.7cm}
\caption {\small Comparing Unidirectional and Bi-Directional Reasoning: Unidirectional reasoning utilizes historical context to make left-to-right decisions. In contrast, bi-directional reasoning explores potential future states and aggregate the expected utility for future moves. It then uses both historical contexts and future exploration contexts to conduct bi-directional reasoning.}
\label{fig
}
\vspace{-0.4cm}
\end{wrapfigure}

BIDDER's approach involves three main processes: inferring hidden states to represent uncertain information from historical data, using these hidden states to explore future potential states and outcomes, and integrating both historical contexts and future explorations to inform reasoning. By leveraging these bi-directional deliberation processes, BIDDER enables LLMs to make more informed and rational decisions. Although future information cannot be directly obtained, BIDDER refers to the Q-Learning \cite{watkins1992q} algorithm in Reinforcement Learning, employing LLMs to explore various future time horizons and use this information to calculate the expected utility of potential actions.

To demonstrate BIDDER's effectiveness, we conducted experiments in two well-defined strategic decision-making scenarios: Poker (Limit Texas Hold'em) \cite{zha2020rlcard} and Negotiation \cite{lewis2017deal, cao2018emergent}. These scenarios provide a rigorous testbed for evaluating the decision-making capabilities of LLMs due to their clear comparison of final payoffs and well-established theoretical optimal solutions. Our experimental results show that BIDDER significantly enhances the decision-making capabilities of LLMs, aligning their behavior more closely with optimal solutions. By incorporating bi-directional deliberation reasoning, BIDDER minimizes uncertainty, considers long-term payoffs, and ensures that LLMs can effectively integrate both historical context and future explorations into their decision-making processes.

The contributions of this work are as follows:
\begin{itemize}
\item We introduce the first \textbf{bi-directional reasoning approach} with LLMs to enhance their decision rationality by considering both historical context and future potential outcomes. Unlike traditional methods that focus solely on past information, this bi-directional reasoning helps manage uncertainty and predict expected utility more effectively.
\item We incorporate principles from the Q-Learning algorithm, leveraging LLMs to \textbf{explore future trajectories and aggregate rewards}, thereby constructing reasoning contexts that include potential future outcomes.
\item We validate BIDDER's effectiveness through rigorous experiments in strategic decision-making scenarios, demonstrating its superiority over traditional reasoning methods.
\end{itemize}

\section{Method}

\begin{figure*}[ht]
\centering
    \vspace{-2mm}
  \includegraphics[width=0.94\linewidth]{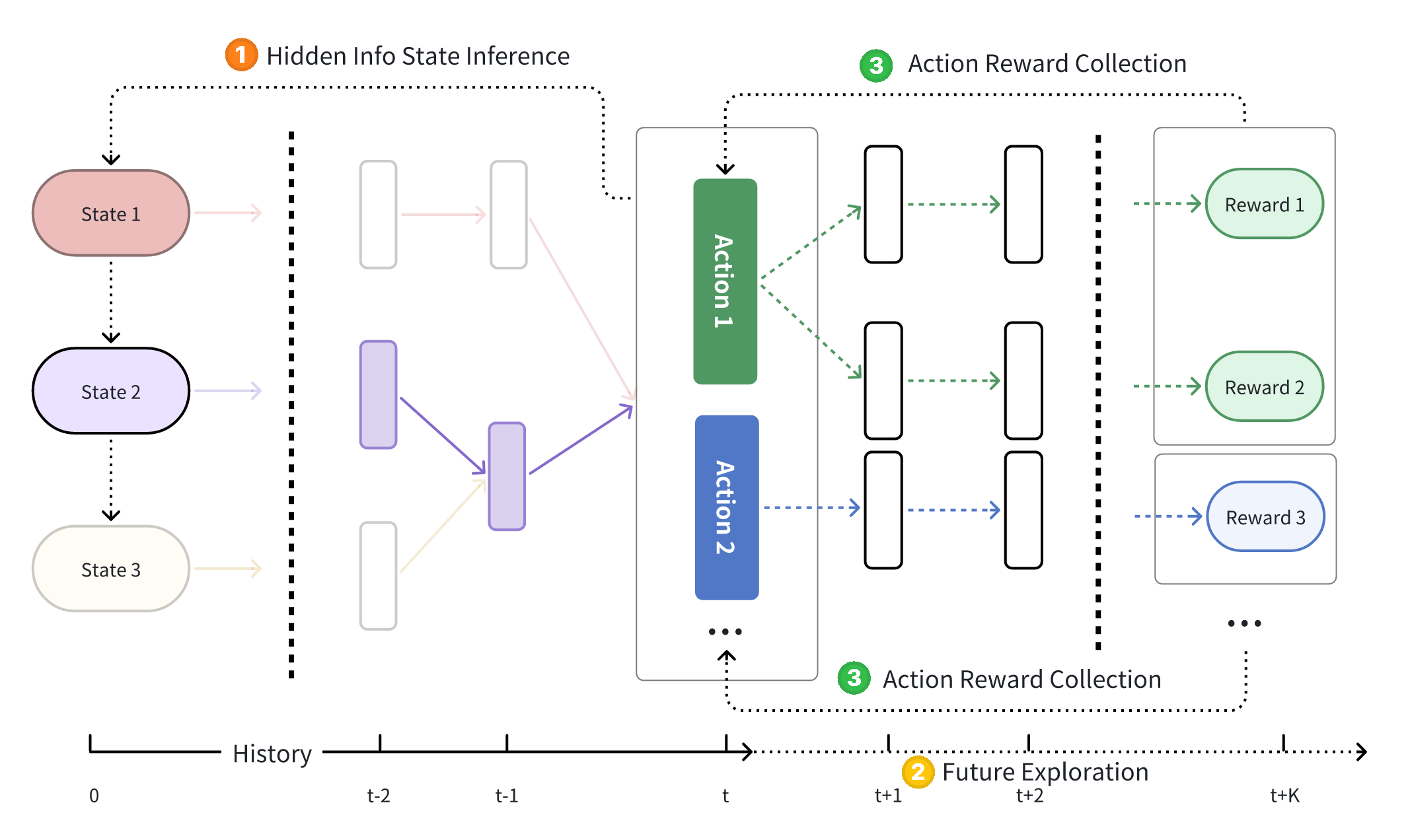} 
  \vspace{-5mm}
  \caption {\textbf{BI}-\textbf{D}irectional \textbf{DE}liberation \textbf{R}easoning includess: 1. Hidden State Inference: Inferring hidden states (e.g. opponent's strategies) behind decision-making from historical data (e.g. opponent's actions); 2. Future Exploration: Exploring possible future trajectories based on inferred states; 3. Action Reward Collection: Aggregating utilities from predicted trajectories to make decisions. The bi-directional reasoning enhances decision rationality by incorporating both historical context and potential future outcomes.}
\label{fig:architecture}
\vspace{-5mm}
\end{figure*}

\textbf{BI}-\textbf{D}irectional \textbf{DE}liberation \textbf{R}easoning (BIDDER) is designed to enhance the decision rationality of LLMs. It is based on the concepts of uncertainty and expected utility in decision theory \cite{Bacchus1996FromSK} and leverages bi-directional reasoning to incorporate both historical context and potential future outcomes into the decision-making process. Since future contexts are not available at the decision point, we refer to the Q-Learning algorithm in Reinforcement Learning \cite{watkins1992q}, calculating the expected total reward for taking different actions in a given state as the context for right-to-left reasoning. 

Figure \ref{fig:architecture} illustrates the entire process of this method. Specifically, BIDDER consists of three modules: \textbf{1)} inferring hidden states behind decision-making from available historical data, \textbf{2)} utilizing the inferred hidden states to explore future trajectories, and \textbf{3)} aggregating the utilities along each explored trajectories and making decisions to maximize the expected long-term utility.

In order to simplify the complexity of the description, we will limit the subsequent discussion to a two-player game, referring to the current player as "player" and the other participant as "opponent."
We standardize the notation here:
\begin{itemize}[left=2mm]
    \item The player's initial hidden state $s_0$
 , and the opponent's hidden state $\overline{s}_0$.
 \item
 The player's public trajectory at time $t$:
$\mathcal{\tau}(t) =\{a_0, s_1, \dots, s_t\}$ (as the hidden state, $s_0$ is unobservable by other players) , and the opponent's trajectory  $\mathcal{\overline{\tau}}(t)$.
\end{itemize}

\subsection{Infer Hidden State From History}
\label{sec.infer}
Inferring hidden states is essential for understanding environmental changes influenced by latent variables, such as an opponent's status or strategy in a game. Predicting these variables allows for a more accurate reconstruction of future behaviors, enhancing the rationality of decision-making.

Previous research \cite{GruverFQW23, XieRL022} has shown that LLMs can recognize contextual causal relationships and correct causal arguments with high probability. 
Therefore, we use a causal reasoning approach to infer environmental information, with historical data $\mathcal{\overline{\tau}}(t)$ as the effect and environmental uncertainties $\overline{s}_0$ as the causes.


Formally, we consider the relationship \(P(\mathcal{\overline{\tau}}|\overline{s}_0)\), where \(P\) represents the probability distribution. By analyzing \(P(\mathcal{\overline{\tau}}|\overline{s}_0)\), we aim to estimate the causal factors \(\overline{s}_0\) influencing the observed data \(\mathcal{\overline{\tau}}\). 
The LLM can be regarded as an implicit Bayesian inference model, inferring hidden states \(\overline{s}_0\) based on observed data \(\mathcal{\overline{\tau}}\), et al. \(P(\overline{s}_0|\mathcal{\overline{\tau}})\).

\begin{wrapfigure}{r}{0.50\textwidth}
\vspace{-6mm}
\begin{mybox}{Hidden State Inference Prompt\\
$\overline{s}_0 = \mathrm{Hidden\_Inference}(\mathcal{\overline{\tau}})$}
\small
Opponent Action History: \vspace{1.5mm}\\
\hl{[Opponent Trajectory]} \vspace{1.5mm}\\
Possible Hidden States for Opponent's Actions:\vspace{1.5mm}\\
\hl{[Discretized State Space List]} \vspace{1.5mm}\\
Select the most likely state of the opponent.
\end{mybox}
\vspace{-5mm}
\end{wrapfigure}


In this prompt, \hl{[Opponent Trajectory]} represents the opponent's observed actions (historical data \(\mathcal{\overline{\tau}}\)), and \hl{[Discretized State Space List]} enumerates possible hidden states \(\overline{s}_0)\) causing these actions. The LLM processes this information and selects the most probable hidden state \(\overline{s}_0\), thereby inferring the causes from the effects. This method leverages the LLM's ability to understand and reason about complex causal relationships, enabling effective prediction and decision-making in uncertain environments.



\subsection{Explore Future Trajectories with Opponent Modeling}
\begin{wrapfigure}{r}{0.50\textwidth}
\vspace{-8mm}
\begin{mybox}{Opponent Modeling Prompt\\
$\overline{a}'_{t} = \mathrm{Opponent\_Modeling}(\overline{\tau}(t))$}
\small
Inferred Hidden State of Opponent:\vspace{1.5mm}\\
\hl{[Opponent's Hidden State]} \vspace{1.5mm}\\
Opponent Action History:\vspace{1.5mm}\\
\hl{[Opponent Trajectory]} \vspace{1.5mm}\\
Select the most likely action which opponent will take under inferred hidden state. 
\end{mybox}
\vspace{-16mm}
\end{wrapfigure}

Based on the hidden states inferred in Section \ref{sec.infer} and opponent's behavioral records, we simulate their likely actions under various states using prompts.

At the current decision point $t$, we explore all possible actions with the following termination conditions:

\begin{itemize}[left=2mm]
\vspace{-1.5mm}
\item Reaching the preset exploration horizon $T$.
\item Reaching a terminal state or a chance node\footnote{A chance node signifies the introduction of new information that can influence decisions. For instance, in Texas Hold'em poker, this occurs when a new community card is revealed.}.
\end{itemize}
For the current reasoning point, we generate a set of exploration paths based on opponent modeling:

\begin{equation}
\begin{array}{c}
\label{eq:Trajectories}
\mathcal{T}(t) = \{\tau \mid \tau = (s_t, a'_t, \ldots, a'_{t+T-1}, s_{t+T})\}
\end{array}
\end{equation}
\begin{equation}
s_{t+1} = s_t\times a'_t \times \overline{a}'_t 
\end{equation}

Here, \(\mathcal{T}(t)\) represents the set of all possible trajectories starting from the current state \(s_t\). Each trajectory \(\tau\) consists of a sequence of states and actions, starting from the current state \(s_t\), and includes the sequence of actions \(a'_t, \ldots, a'_{t+T-1}\) and the explored states \(s_{t+1}, \ldots, s_{t+T}\) up to the exploration horizon \(T\). 
For the detailed algorithm of future trajectories exploration, please refer to Appendix \ref{app:algorithm}.

\subsection{Aggregate Rewards for Explored Trajectories}
\begin{wrapfigure}{r}{0.50\textwidth}
\vspace{-6mm}
\begin{mybox}{Action Reward $r$ Calculation Prompt\\
$r(s_t) = \mathrm{Reward\_Calculation}(\overline{s}_0,s_t)$}
\small
All historical information (including Player and Opponent):\vspace{1.5mm}\\
\hl{[Trajectories of Player and Opponent]} \vspace{1.5mm}\\
Feasible Actions of the Current State:\vspace{1.5mm}\\
\hl{[Feasible Actions]} \vspace{1.5mm}\\
Estimate the possible payoff $r$ for each action, and $r$ must be normalized to between 0 and 1.
\end{mybox}
\vspace{-8mm}
\end{wrapfigure}

We use LLMs to estimate the rewards for the potential payoff of explored paths. This extends the paths in Equation \ref{eq:Trajectories} to include estimated rewards. We utilize the Bellman equation \cite{bellman1966dynamic, sutton2018reinforcement} to aggregate the return of each trace $\tau$:
\begin{equation}
\label{eq:reward}
G(\tau) = \sum_{k=0}^{T} \beta^{k} r(s_{t+k})
\end{equation}

Here, \(G(\tau)\) is the total reward accumulated along the path \(\tau\), \(r(s_{t+k})\) is the reward at horizon \(k\), and \(\beta\) is the discount factor, determining the importance of future rewards.



The return for each current action \(a\) is calculated as follows:
\begin{equation}
\begin{array}{c}
G(a) = \max_{\tau \in \mathcal{T}(t, a)} G(\tau)
\end{array}
\end{equation}
\begin{equation}
    \mathcal{T}(t, a) = \{\tau \mid \tau = (s_t, a'_t=a, \ldots, a'_{t+T-1}, s_{t+T})\}
\end{equation}

Here, \(G(a)\) is the maximum cumulative return starting from the current time step \(t\) and taking action \(a\). The set \(\mathcal{T}(t, a)\) represents all possible trajectories starting from the current state \(s_t\) and action \(a\).

Finally, the LLM makes decisions based on the past available information, the inferred hidden states and the future long-term returns of each action at the current point in time:

\begin{mybox}{Decision Making Based on Bidirectional Deliberation Reasoning\\
$a_t^* = \mathrm{Decision\_Making}(s_t, \overline{s}_0, \{G(a)\})$}

\small
\hl{[Past Trajectory and Player's Current State]} \vspace{1.5mm}\\
Inferred Hidden State of Opponent:\vspace{1.5mm}\\
\hl{[Opponent's Hidden State]} \space\#Left-to-right Deliberation Reasoning  \vspace{1.5mm}\\
Possible Hidden States of Opponent:\vspace{1.5mm}\\
\hl{[$\{G(a)\}$ of all feasible actions in current state]}\space\#Right-to-left Deliberation Reasoning \vspace{1.5mm}\\
Make the most rational decisions based on current information.
\end{mybox}

This bidirectional deliberation reasoning framework enhances decision rationality by incorporating both historical contexts and potential future outcomes into the decision-making process to minimize uncertainty.

\section{Experiments}
\subsection{Limit Texas Hold'em}
\label{sec:poker}
We utilized two-player Limit Texas Hold'em as our testbed. Limit Texas Hold'em is a popular betting game which, unlike no-limit Texas Hold'em, restricts the number of raises and fixes the raise amount. The detiled game rule of Limit Texas Hold'em can be found in the Appendix \ref{app: setting_poker}. Among the two game participants, we refer to the focal participant as the Player, while the other participant is designated as the Opponent.

We randomly generated 100 game sessions. To enhance the diversity of the hands, we increased the frequency of strong hands appearing in these 100 games. The specific distribution of hand types is detailed in the Appendix\ref{app: setting_poker}. For fairness, each game has a mirrored game, where the Player and the Opponent switch their hands.
\subsubsection{Baselines}
\label{sec:poker_baseline}
We adapt a variety of approaches to comprehensively compare the performance of methods from different sources in Limit Texas Hold'em environment. These methods include:

\textbf{Rule-Based Approaches}: 

\textit{Random}: Randomly selects an action from the set of feasible actions.

\textit{Rule}: Predefined procedures that select actions based on hand strength, such as raising when holding a pair.

\textit{Counterfactual Regret Minimization (CFR)} \cite{zinkevich2007regret} : An iterative algorithm that converges to a Nash equilibrium in two-player zero-sum games.

\textbf{Machine Learning Approaches}:

\textit{Deep Q-Network (DQN)} \cite{mnih2015human}: Utilizes deep neural networks to approximate the Q-value function, addressing decision-making problems in high-dimensional state spaces.

\textit{Deep Monte Carlo (DMC)} \cite{silver2016mastering}: Integrates deep learning with Monte Carlo methods, using deep neural networks to generate efficient samples to estimate the expected value of complex distributions.

\textit{Deep CFR} \cite{brown2019deep}: Combines deep learning with counterfactual regret minimization, employing neural networks to approximate regression targets for strategy solving in complex games.

\textbf{LLM Reasoning Approaches}:

\textit{Direct}: The LLM generates the final action in response to the given game setting prompt.

\textit{Chain-of-Thought (CoT)} \cite{wei2022chain, kojima2022large}: A zero-shot native reasoning method where the LLM follows a step-by-step thought process.

\textit{Reflexion} \cite{shinn2023reflexion}: This method refers to the concept of language agents with verbal reinforcement learning.

\subsubsection{Main Result}

We utilized the gaming environment described in the Section \ref{sec:poker}. Table \ref{tab:payoff} presents the performance of different methods when facing various opponents, with the values representing the average final chips obtained over 100 games.

\input{tabel/tab_payoff}

From Table \ref{tab:payoff}, it is evident that compared to the previous LLM Reasoning method (Direct, CoT and Reflexion), BIDDER shows significant improvement when facing different opponents, demonstrating a noticeable increase in average earnings.

Additionally, in comparison with Machine Learning methods, BIDDER astonishingly outperformed. It is noteworthy that BIDDER relies solely on LLM for bidirectional information judgment, operating as a completely non-learning method. This highlights the potential of BIDDER in decision-making based on LLM.

\subsubsection{Rationalality Evaluation}

\input{tabel/tab_nash}
Equilibrium can be viewed as a stable state in a multi-agent game environment, achieved by each participant through a rational decision-making process. In most cases, it can be regarded as the most rational decision.
We take the behavior at each decision point of DeepCFR as the optimal solution $a_o$, and compare the rationality levels of different methods with the optimal solution. 

The quantitative calculation method is:

\begin{equation}
\label{eq:rational}
    \mathrm{Rational\ Degree} = \frac{\#\{a=a_o \}}{\#\{\mathrm{All\ action}\  a\}}
\end{equation}

\begin{wrapfigure}{r}{0.5\textwidth}
\vspace{-0mm}
\includegraphics[width=1\linewidth]{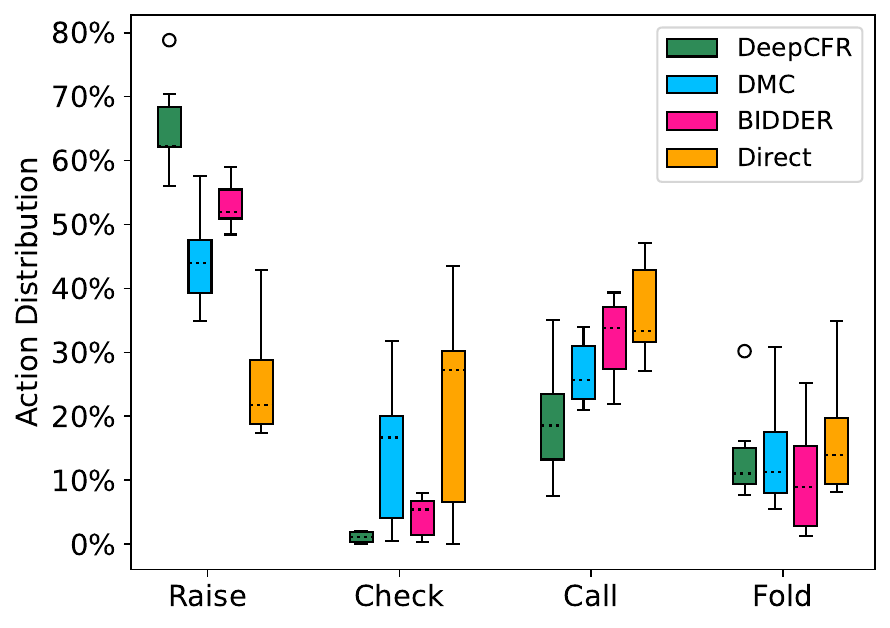}
\vspace{-7mm}
\label{fig:dist}
\caption {\small Direct, BIDDER, and DeepCFR Action Distribution in Limit Texas Hold'em.}
\vspace{-5mm}
\end{wrapfigure}

Table 2 shows the rationality levels of different methods when facing various opponents. From Table 2, we can see that BIDDER has significantly improved the $\mathrm{Rational\ Degree}$ compared to previous LLM Reasoning methods and Machine Learning methods. This indicates that BIDDER has enhanced the decision rationality of the LLM.

Additionally, we further analyzed the action distribution of different methods in Limit Texas Hold'em. From Figure 3, we can observe that compared to the Direct approach, BIDDER has a much higher frequency of Raise, which aligns with the optimal solution (DeepCFR). In contrast, the Direct approach shows higher frequencies of Call and Check, indicating that an LLM-based approach acts as a passive decision-maker. BIDDER, however, calculates future returns and prompts the LLM to act based on these returns, transforming the LLM into an active decision-maker. This active decision-making is rational, leading to improved returns.

\subsection{Negotiation}

Negotiation requires complex communication and reasoning skills, making it an ideal environment for studying artificial intelligence and classical game theory \cite{nash1950equilibrium, von1947theory} due to its dynamic, controllable, and easily assessable nature. We utilize a game paradigm known as "division of a common pool of items," as defined in previous research. In this game, agents are tasked with establishing a mutually acceptable division of items while having their own hidden utilities for each item. Our rules are based on those outlined by \citeauthor{cao2018emergent}, with the following modifications: ensuring that each individual's total value for all items is equal (consistent with \citeauthor{lewis2017deal}), which is more equitable compared to the original uniform distribution between 1 and 10.  The experiment consists of 100 games, with each game having a different number and distribution of item values. Additionally, symmetrical experiments are conducted by swapping the values for both parties in another game.
\subsubsection{Baselines}
We use widely adopted baselines for reasoning and decision-making comparisons, including Direct and CoT introduced in Section \ref{sec:poker_baseline}. Additionally, we consider:

\textit{Monte Carlo Tree Search (MCTS)} \cite{chaslot2008monte}: This is a heuristic search algorithm that efficiently explores large search spaces by sampling potential future outcomes through Monte Carlo simulations.

\textit{Tree of Thoughts (ToT)} \cite{yao2023tree}: This approach simulates human System 2 thinking and excels in complex problems, particularly those requiring search and judgment.

It is important to note here that although BIDDER fundamentally employs tree search for future exploration, similar to MCTS and ToT, BIDDER's search process integrates opponent hidden information inference and opponent behavior modeling. This contrasts with MCTS's probabilistic sampling and ToT's self-exploration approach.
\subsubsection{Metrics}

We employ three metrics for negotiation as defined by \citeauthor{lewis2017deal}, which are:

\textbf{Score}: The average score obtained in each game where an agreement is reached. The score is calculated as the sum of the products of each item assigned to the agent and its value.

\textbf{Agreement}: The probability of reaching an agreement across all games.

\textbf{Pareto Optimality}: The percentage of Pareto optimal solutions for the agreed deals. \footnote{Pareto optimal refers to a state in which it is impossible to improve individual's situation without worsening other's.}

These metrics not only reflect the agent's competitive advantage in negotiations but also demonstrate its cooperative behavior.

\subsubsection{Result}

\input{tabel/tab_neg}

From Table \ref{tab:neg}, we can observe that compared to MCTS, the reasoning methods based on LLM show significant improvements over the Direct approach in both Score and Agreement metrics. Additionally, all methods achieved 100\% Agreement when competing against MCTS. This is because MCTS is more inclined to agree, yet it fails to achieve higher gains in terms of score.  Furthermore, compared to the results reported by \citeauthor{lewis2017deal}, the scores for LLM reasoning methods on Pareto optimality are generally lower. We believe there are two possible reasons for this: first, the tested methods are all zero-shot, so LLMs cannot effectively predict the timing for achieving Agreement; second, negotiation is a game that requires both competition and cooperation, and balancing these two aspects poses a challenge for LLMs.

Moreover, we found that BIDDER achieved moderate Agreement but had the highest Score (payoff). This also reflects BIDDER's rational decision-making when aiming for payoff maximization, especially when long-term planning is required.

\section{Related Work}
\subsection{Enhanced Reasoning Methods of LLMs}
Large Language Models (LLMs) demonstrate remarkable proficiency in a range of complex reasoning tasks, such as mathematical reasoning \cite{miao2021diverse, patel2021nlp}, common sense understanding \cite{talmor2022commonsenseqa, bhakthavatsalam2021think}, and symbolic reasoning \cite{srivastava2022beyond, suzgun2022challenging}. A key technique in enhancing LLM reasoning is the Chain-of-Thought (CoT) method, which involves decomposing complex questions into a series of intermediate steps \cite{wei2022chain, kojima2022large}. Building upon CoT, researchers have developed methods such as Tree of Thought (ToT) \cite{yao2023tree}, Graph of Thought (GoT) \cite{besta2023graph}, and Skeleton-of-Thought \cite{ning2023skeleton} to further advance reasoning processes.
To enhance the consistency and quality of LLM responses, methods like Self-Refine \cite{madaan2023self} and Reflexion \cite{shinn2023reflexion} introduce iterative refinement, while several works \cite{fu2023improving, wang2023unleashing} integrate detailed persona information to improve rationality and knowledge precision in LLM reasoning.


Existing methods such as ToT and GoT enhance reasoning performance by navigating through the problem space and selecting optimal nodes (states). States are evaluated based on the progress toward solving the problem. However, these approaches typically involve unidirectional state transitions and reward generation. In contrast, our work introduces a bidirectional approach, allowing for a more dynamic and rational reasoning process.

\subsection{Heuristic Search for LLM Reasoning}
With the widespread application of LLMs in the field of natural language processing, enhancing their reasoning capabilities has become a research focus. Heuristic search algorithms, as an effective optimization technique, are being introduced into LLM reasoning to improve efficiency and accuracy. Specifically,  \citeauthor{yao2023tree} proposed the ToT algorithm based on "System2," which constructs a search tree by generating candidate actions through LLM and evaluating them. Subsequently, more methods such as A* algorithm and Monte Carlo Tree Search (MCTS) have been proposed to enhance LLM's reasoning and planning capabilities \cite{feng2023alphazero, hao2023reasoning, zhuang2023toolchain}. To further improve the efficiency and accuracy of search paths, researchers have also introduced major voting\cite{wang2022self}, self-verification\cite{weng2022large,zhang2023controlling}, training value functions\cite{wang2024q}, and game-theory-based solutions\cite{gemp2024states, 
 gandhi2023strategic}.
 
However, these methods are primarily used for reasoning tasks. In multi-agent decision environments where opponent state information and rewards are uncertain, traditional heuristic search algorithms are insufficient. BIDDER extends the application of heuristic search in LLMs by utilizing historical information to infer hidden opponent information, enabling opponent modeling. This approach allows for exploration and long-term reward calculation in multi-agent environments, thus achieving rational decision-making.

\subsection{Decision-making based on LLM Agents}
Recent research on large language models (LLMs) in decision-making \cite{zhang2024llm} has spanned various multi-agent systems (MAS) including social behavior\cite{zhou2023sotopia, hua2023war}, economic simulations\cite{zhao2023competeai, li2023tradinggpt}, game theory\cite{duan2024gtbench, xu2023magic}, and game playing\cite{ma2023large, xu2023exploring}. 
To enhance the strategic reasoning capabilities of LLMs, researchers have incorporated the concepts of Theory of Mind \cite{gandhi2023strategic, guo2023suspicion, zhang2024k}  and Reinforcement Learning \cite{xu2023language, zhang2024agent}. These approaches aim to prompt LLMs to grasp the complexities of decision-making tasks.

Although these studies have demonstrated the potential and applications of LLMs in the decision-making domain, specialized reasoning frameworks designed for specific tasks face significant challenges when applied to other fields. Our generalized rational decision-making theory introduces a bi-directional deliberation reasoning method. Through various experiments, it has been validated that this approach significantly enhances the benefits gained by LLMs in decision-making processes. 

\section{Conclusions}
As large language models (LLMs) are increasingly applied across various fields,, BIDDER offers a novel approach to enhance their performance in complex decision-making environments. By introducing bidirectional reasoning, BIDDER integrates key concepts from decision science, such as uncertainty management and expected utility, into the reasoning processes of LLMs.
By leveraging both historical data and potential future rewards, BIDDER effectively addresses the limitations of traditional unidirectional reasoning methods. The integration of decision theory principles and bi-directional modeling ensures a comprehensive exploration of possible states, leading to more informed and rational decisions. Our experiments in poker and negotiation scenarios demonstrate that BIDDER significantly improves the decision-making capabilities of LLMs, aligning their performance more closely with theoretical optimal solutions. This advancement represents a significant step forward in the development of LLMs, enabling them to make better decisions in complex, real-world scenarios.

\bibliographystyle{plainnat}
\bibliography{custom}

\appendix
\section{Game Setting}
\subsection{Limit Texas Hold'em}
\label{app: setting_poker}
Limit Texas Hold'em is a variant of Texas Hold'em poker where the betting limits are fixed. The fixed betting structure and sequential decision-making process create a controlled yet challenging environment for researching agent decision-making. Its partial observability, and stochastic nature make it ideal for studying how agents handle uncertainty, strategic planning, and adversarial competition.

Rules and Gameplay are as following:
\begin{enumerate}
    \item Blinds: - Small Blind and Big Blind are posted before the cards are dealt.
   - The Small Blind is usually half the size of the Big Blind.
    \item Betting Rounds:
   - There are four betting rounds: Pre-Flop, Flop, Turn, and River.
   - Betting amounts are fixed and structured according to the round.
   \item Dealing:
   - Each player is dealt two private cards (hole cards).
   - Five community cards are dealt in stages: the Flop (three cards), the Turn (one card), and the River (one card).
   \item Action Spaces: there are four actions, Call, Raise, Fold and Check. In Limit Texas Hold’em. Each player can only choose a fixed amount of raise. And in each round the number of raises is limited to four times.   
\end{enumerate}

For utilizing LLM as a player, the game setting prompt and the state description prompt required for each action are as follows:
\begin{mybox}{Game Setting Prompt}

You are \{name\} and you are a Texas Hold’em player. \\
Texas Hold’em is a popular betting game.  \\
Each player is dealt two face-down cards, called hole cards.  \\
Then 5 community cards are dealt in three stages (the flop, the turn and the river).  \\
Each player seeks the five best cards among the hole cards and community cards.  \\
In this Texas Hold’em game, there are a total of \{num\_players\} players namely \{players\}. \\
The number of chips every player has is infinite. \\
You just need to win more chips in the competition as much as possible.
\end{mybox}

\begin{mybox}{State Prompt}
The current state is:

1. The Public Card is \{public\_card\}, and Your Hand is \{hand\}.

2. You have bet \{my\_bid\} chip, and the number of chips all players have invested is \{all\_chips\}.

3. After your \{pre\_action\}, \{oppoent\_actions\}.

You can choose one of the following actions now: \{actions\}.

\end{mybox}

\begin{figure}[ht]
\centering
\vspace{-5mm}
\includegraphics[width=0.4\linewidth]{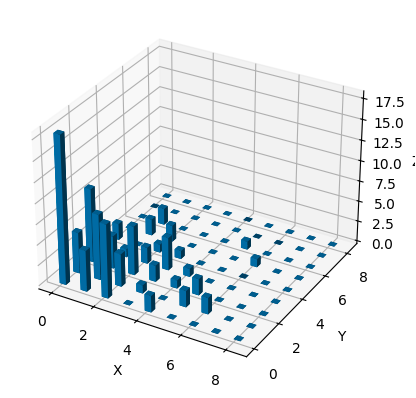}
\vspace{-4mm}
\caption {The distribution of hand strengths sampled from poker games ranges from 0 to 8, corresponding respectively to High Card, One Pair, Two Pair, Three of a Kind, Straight, Flush, Full House, Four of a Kind, and Straight Flush.}
\label{fig:dist_sample}
\vspace{-0.3cm}
\end{figure}

\subsection{Negotiation}

The "Negotiation" game, as described in \citeauthor{cao2018emergent}, is a semi-cooperative model of agent interaction where communication plays a pivotal role. Here are the key rules of the Negotiation game:

\begin{enumerate}[left=2mm]
    \item Item Pool: Agents are presented with a pool of items, which in the paper's example includes three types of items: peppers, cherries, and strawberries. Each type of item has a quantity associated with it, which is sampled uniformly for each round of the game.

    \item Utility Functions: Each agent receives a utility function that specifies the rewarding value of each item. These utilities are private, meaning that each agent only knows its own utility values, which are sampled uniformly with at least one item having non-zero utility.

    \item Negotiation Rounds: Over a series of negotiation timesteps (N), the agents take turns to propose how to divide the items in the pool. The number of turns, N, is sampled between 4 and 10 in each round to prevent a 'first-mover' advantage.

    \item Alternating Turns: The agents alternate turns, with one agent (referred to as Agent A) always acting on odd turns and the other agent (Agent B) on even turns.

    \item Proposals and Messages: During each turn, an agent can make a proposal regarding the division of the items and send a message through a communication channel. The proposal is a vector indicating the quantity of each item type that the agent is requesting.

    \item Termination: Either agent can terminate the negotiation at any timestep with a special action, which signifies agreement with the most recent proposal made by the other agent. The negotiation ends, and rewards are distributed based on the last proposal.

    \item Rewards: If the negotiation is successfully terminated, the agents receive rewards based on the dot product of their utility vectors and the items they receive. If an agent makes an invalid proposal and it is accepted, both agents receive a reward of zero.

    \item No Agreement Scenario: If the negotiation reaches the upper limit of allowed turns without agreement, both agents receive no reward.
\end{enumerate}

The game is designed to study how agents can learn to communicate and negotiate effectively, with a focus on the emergence of communication protocols and strategies in a multi-agent reinforcement learning setting.

For utilizing LLM as a player, the game setting prompt and the state description prompt required for each action are as follows:
\begin{mybox}{Game Setting Prompt}
You are negotiating the division of Peppers, Strawberries, and Cherries with the opponent. Different values these items hold for both you and your opponent. The process is structured into two stages per round: the proposal stage and the utterance stage.

Remember, the key in such negotiations is understanding that your opponent also has their value system for these items, which is unknown to you. Balancing between revealing your true desires and misleading your opponent to gain a favorable outcome is essential. It\'s also important to be adaptive, as the negotiation progresses and you gather more information about your opponent\'s preferences and tactics.
\end{mybox}

\begin{mybox}{Proposals State Prompt}
Now, you are in the Proposal stage: you\'ll determine the division of items you desire. This is expressed as [a, b, c], where \'a\' represents the quantity of Peppers, \'b\' the quantity of Strawberries, and \'c\' the quantity of Cherries you wish to acquire. It\'s crucial to base this division on the perceived value these items have for you, keeping in mind that the goal is to reach a mutually agreeable solution.
\end{mybox}

\begin{mybox}{Utterance State Prompt}
Now, you are in the Utterance Stage: you communicate to your opponent what you want, again in the format [a, b, c]. This utterance is your strategic communication and doesn\'t necessarily have to reflect your actual desires or the proposal you formulated in the first stage. It\'s a tool for negotiation, potentially used to mislead, bluff, or strategically reveal information to your opponent.
\end{mybox}

\section{Reasoning Method Implementation Details}
\subsection{Limit Texas Hold'em}
\subsubsection{Direct}
\begin{mybox}{Query Prompt}
Please provide your results in the form of \{'action': ''\}. Just output the dictionary, don't use any other text.
\end{mybox}

\subsubsection{CoT}
\begin{mybox}{Query Prompt}
Please first think and reason about the current state and then generate your action as follows: 
My thought is \{Your Thought\}, and my action is \{'action': ''\}
\end{mybox}

\subsubsection{Reflexion}
\begin{mybox}{Query Prompt}
Please first think and reason about the current state and then generate your action as follows: 
My thought is \{Your Thought\}, and my action is \{'action': ''\}
\end{mybox}

\begin{mybox}{Reflexion Prompt}
Please carefully check the thought and the action you just output , and then refine your answer. The final output is also in the same format: 
My revised thought is \{Your Thought\}. My revised action is \{'action: ''\}.
\end{mybox}

\subsubsection{BIDDER}
\begin{mybox}{Hidden State Inference Prompt}
You are a Poker expert. Based on the public information provided below, evaluate the relative strength of each player's hand. The strongest hand should be rated as 5, and the weakest as 1.\\
Available History Information:\\
\{events\}\\
Instructions:\\
1. Evaluate the relative strength of each player's hand based on the information given.\\
2. Consider the players' actions and the community cards in your evaluation.\\
3. Provide a rating from 1 to 5, with 5 being the strongest hand and 1 being the weakest\\
Please provide your evaluation based on the information above.\\
Sample Output Format:\\
\{\\
"Player": \{{\\
"Rating": <rating>\\
\}}\\
\}\\
\end{mybox}

\begin{mybox}{Modeling Prompt}
You are observing an ongoing poker game. Here are the details:

- Current community cards: \{public\_cards\}
- Players and their estimated hand strengths:
  \{hand\_strength\}

The game history so far is as follows:
{events}

What is the most likely action for {player} when legal actions are: \{legal\_action\}?

The JSON format should look like this:
\{"action": "<most likely action>"\}
\end{mybox}

\begin{mybox}{Reward Prompt}
You are observing an ongoing poker game. Here are the details:\\
- Current community cards: \{public\_cards\}\\
- Players and their estimated hand strengths:\\
  \{hand\_strength\}\\
The game history so far is as follows:\\
{events}\\
Predict the potential payoff (normalized between 0 and 1) for \{player\} for each of the following possible actions and return the results in JSON format:
\{action\_payoff\_template\}
\end{mybox}

\subsection{Negotiation}
For the LLM reasoning baselines in the Negotiation game (including Direct, CoT, and ToT), we utilize the implementation provided by \citeauthor{duan2024gtbench}\footnote{https://github.com/jinhaoduan/GTBench}.

\subsubsection{BIDDER}
\begin{mybox}{Hidden State Inference Prompt}
\{state\}\\
Based on the public information provided below, infer the value of each item to your opponent.\\

Opponent behavior:\\
\{events\}  \\
Item Pool:\\
\{item\_pool\_prompt\}\\
Instructions:\\
1. Consider the players' historical actions in your evaluation.\\
2. The value is between 1 and 10.\\
Please provide your evaluation based on the information above.\\
Output Format:\\
Evaluation: \\
Peppers: <value> \\
Strawberries: <value>  \\
Cherries: <value> 
\end{mybox}

\begin{mybox}{Reward Prompt}
You are negotiating the division of Peppers, Strawberries, and Cherries with the opponent. Different values these items hold for both you and your opponent. The process is structured into two stages per round: the proposal stage and the utterance stage.\\
Now, you are in the Utterance Stage: you communicate to your opponent what you want, again in the format [a, b, c]. This utterance is your strategic communication and doesn't necessarily have to reflect your actual desires or the proposal you formulated in the first stage. It's a tool for negotiation, potentially used to mislead, bluff, or strategically reveal information to your opponent.\\
\{item\_pool\_prompt\}\\
\{value\_vector\}\\
\{last\_utterance\_prompt\}\\
The opponent's evaluation of the value of each item is:\\
Opponent Value Inference: \\
\{infered\_values\}\\
Instructions:\\
1. Propose the two Utterances with the largest rewards. This also includes agreeing if you agree with your opponent's utterance. Each utterance format is: "<Utterance: [a, b, c]>" (including "<" and ">").\\
2. Calculate the Payoff for each Utterance (a number); calculate the probability (normalized between 0 and 1) of the opponent agreeing with this utterance, it is necessary to consider the degree of satisfaction the opponent has with the utterance.\\

Sample Output Format:\\
Utterances:\\
\{\\
    "Utterance": "<utterance>",\\
    "Payoff": "<payoff>",\\
    "Probability": "<probability>"\\
\},\\
\{\\
    "Utterance": "<utterance>",\\
    "Payoff": "<payoff>",\\
    "Probability": "<probability>"\\
\}
\end{mybox}

\section{Algorithm Detail of BIDDER}
\label{app:algorithm}

\begin{algorithm}[ht]
    \caption{\textbf{Explore} $(\mathcal{T},t)$: Exploration with Opponent Modeling}
    \label{alg:explora}
    \begin{algorithmic}[1]
    \REQUIRE $s_t$: Current decision context; 
    $I$: Inferred hidden information about opponent; 
    $T$: Time-step of exploration; 
    \ENSURE $\mathcal{T}(s_t)$: The traces of exploration.
         \IF{$t = T$ or $s_t$ is terminal node}  
            \RETURN $\{\tau\}$
        \ENDIF
        \STATE $\mathcal{T} = \emptyset $ 
        \FOR{each legal action $a_t$ of $s_t$}
            \STATE $s_{t+1} \gets   s_t \times a_t \times$ {\textbf{Opponent\_Modeling($I$, $\tau$)}}
            \STATE $\mathcal{T} \gets \mathcal{T}  \cup $   \textbf{Explore}$(\tau \cup \{a_t, s_{t+1}\}, t+1)$
        \ENDFOR
        \RETURN $\mathcal{T}$
    \end{algorithmic}
\end{algorithm}

\begin{algorithm}[ht]
    \caption{Rewards Aggregation with Attenuation Coefficient}
    \label{alg:aggregate}
    \begin{algorithmic}[1]
    \REQUIRE $\mathcal{T}$: Explored trajectories; 
    $A$: The set of valid actions; 
    $\beta$: Attenuation Coefficient; 
    \ENSURE $G(A)$: The maximum payoff of each action.

     \FOR{each $a \in A$}
        \STATE $G(a) \gets 0$
        \FOR{each trajectory $\tau \in \mathcal{T}$}
            \IF{$\tau$ starts with action $a$}
                \STATE $G(\tau) \gets \sum_{k=0}^{T} \beta^i \cdot \tau[k]$
                \IF{$G(\tau) > G(a)$}
                    \STATE $G(a) \gets G(\tau)$
                \ENDIF
            \ENDIF
        \ENDFOR
    \ENDFOR

    \STATE \RETURN $\{G(a) \mid a \in A\}$
    \end{algorithmic}
\end{algorithm}

\end{document}

%% file: settings.tex
\usepackage{multirow}
\usepackage{amsmath}
\usepackage{capt-of}
\usepackage{tabularx}
\usepackage{epsfig}
\usepackage{amssymb}
\usepackage{amsfonts}
\usepackage{booktabs}
\usepackage{scalerel}
\usepackage[inline]{enumitem}
\usepackage{listings}
\usepackage{varwidth}
\usepackage[export]{adjustbox}
\usepackage{tikz}
\usetikzlibrary{tikzmark}

\usepackage{stmaryrd}
\usepackage{bbm}
\usepackage{wrapfig}
\usepackage{pifont}
\usepackage[noabbrev]{cleveref}

\definecolor{deepblue}{rgb}{0,0,0.5}
\definecolor{officeblue}{RGB}{0,102,204}
\definecolor{deepred}{rgb}{0.6,0,0}
\definecolor{deepgreen}{rgb}{0,0.5,0}
\definecolor{mybrickred}{RGB}{182,50,28}
\definecolor{fillcolor}{RGB}{216,217,252}

\usepackage{etoolbox}
\usepackage{framed}

\newif\ifxetexorluatex
\ifxetex
  \xetexorluatextrue
\else
  \ifluatex
    \xetexorluatextrue
  \else
    \xetexorluatexfalse
  \fi
\fi
\newcommand*\quotesize{60} 
\newcommand*{\openquote}
   {\tikz[remember picture,overlay,xshift=-4ex,yshift=-2.5ex]
   \node (OQ) {\fontsize{\quotesize}{\quotesize}\selectfont``};\kern0pt}

\newcommand*{\closequote}[1]
  {\tikz[remember picture,overlay,xshift=4ex,yshift={#1}]
   \node (CQ) {\fontsize{\quotesize}{\quotesize}\selectfont''};}

\colorlet{shadecolor}{white}

\newcommand*\shadedauthorformat{\emph} 

\newcommand*\authoralign[1]{%
  \if#1l
    \def\authorfill{}\def\quotefill{\hfill}
  \else
    \if#1r
      \def\authorfill{\hfill}\def\quotefill{}
    \else
      \if#1c
        \gdef\authorfill{\hfill}\def\quotefill{\hfill}
      \else\typeout{Invalid option}
      \fi
    \fi
  \fi}
{\authoralign{#1}
\ifblank{#2}
   {\def\shadequoteauthor{}\def\yshift{-2ex}\def\quotefill{\hfill}}
   {\def\shadequoteauthor{\par\authorfill\shadedauthorformat{#2}}\def\yshift{2ex}}
\begin{snugshade}\begin{quote}\openquote}
{\shadequoteauthor\quotefill\closequote{\yshift}\end{quote}\end{snugshade}}

\newfloat{Code}{htbp}{Code}

%% file: math_commands.tex
\usepackage{amsmath,amsfonts,bm}

\def\eqref#1{equation~(\ref{#1})}

\def\1{\bm{1}}

\DeclareMathAlphabet{\mathsfit}{\encodingdefault}{\sfdefault}{m}{sl}
\SetMathAlphabet{\mathsfit}{bold}{\encodingdefault}{\sfdefault}{bx}{n}



%% file: tabel/tab_payoff.tex
\begin{table*}[ht]
    \centering
    \resizebox{0.8\textwidth}{!}{
    \begin{tabular}{@{}l|l|cccccc|c@{}}
    \toprule
            & \textbf{\begin{tabular}[c]{@{}l@{}}\diagbox{\textbf{Player}}{\hspace{-1cm}\textit{Opponent}}\end{tabular}} & \multicolumn{1}{l}{\textit{Random}} & \multicolumn{1}{l}{\textit{Rule}} & \multicolumn{1}{l}{\textit{CFR}} & \multicolumn{1}{l}{\textit{DQN}} & \multicolumn{1}{l}{\textit{DMC}}  & \multicolumn{1}{l}{\textit{DeepCFR}} & AVG                                   \\ \midrule
            & \textbf{Random}                                                     & \cellcolor[HTML]{FFE0E0}-0.28       & \cellcolor[HTML]{FFEAEA}-0.05 & \cellcolor[HTML]{FFD3D3}-0.56     & \cellcolor[HTML]{FF9898}-1.87    & \cellcolor[HTML]{FFD4D4}-0.55       & \cellcolor[HTML]{FF9898}-1.89        & \cellcolor[HTML]{FFC5C5}-0.87         \\
        \multirow{-3}{*}{Rule}                                                                 & \textbf{Rule}                                                       & \cellcolor[HTML]{FFEDED}0.01        & \cellcolor[HTML]{FFE6E6}-0.14 & \cellcolor[HTML]{FFEEEE}0.04     & \cellcolor[HTML]{FFF6F6}0.21     & \cellcolor[HTML]{FFE9E9}-0.07         & \cellcolor[HTML]{FFFBFB}0.33         & \cellcolor[HTML]{FFEFEF}0.06          \\
            & \textbf{CFR}                                                        & \cellcolor[HTML]{F6FBF4}0.55        & \cellcolor[HTML]{FFE5E5}-0.17 & \cellcolor[HTML]{FFDBDB}-0.38    & \cellcolor[HTML]{FF5F5F}-3.15    & \cellcolor[HTML]{FF9898}-1.89        & \cellcolor[HTML]{FF3E3E}-3.88        & \cellcolor[HTML]{FFAAAA}-1.49         \\ 
 \midrule
            & \textbf{DQN}                                                        & \cellcolor[HTML]{71B958}2.61        & \cellcolor[HTML]{FFF5F5}0.19    & \cellcolor[HTML]{61B145}2.86    & \cellcolor[HTML]{FFEFEF}0.06     & \cellcolor[HTML]{FFEBEB}-0.03       & \cellcolor[HTML]{FF0000}-5.28        & \cellcolor[HTML]{FFEFEF}0.07          \\
            & \textbf{DMC}                                                        & \cellcolor[HTML]{B7DBAA}1.53        & \cellcolor[HTML]{FFFCFC}0.34  & \cellcolor[HTML]{BFDFB3}1.41     & \cellcolor[HTML]{FFF4F4}0.17     & \cellcolor[HTML]{FFE8E8}-0.09        & \cellcolor[HTML]{FFEBEB}-0.038       & \cellcolor[HTML]{F6FBF4}0.55          \\
        \multirow{-3}{*}{\begin{tabular}[c]{@{}l@{}}Reinforcement \\  Learning\end{tabular}}   & \textbf{DeepCFR}                                                    & \cellcolor[HTML]{B6DBA8}1.55        & \cellcolor[HTML]{FFF1F1}0.11  & \cellcolor[HTML]{62B145}2.85    & \cellcolor[HTML]{4EA72E}3.15     & \cellcolor[HTML]{FFEBEB}-0.03         & \cellcolor[HTML]{FFCECE}-0.67        & \cellcolor[HTML]{CFE7C6}1.16          \\ \midrule
            & \textbf{Direct}                                                     & \cellcolor[HTML]{E2F1DD}0.86        & \cellcolor[HTML]{EEF7EB}0.68   & \cellcolor[HTML]{EAF5E6}0.74     & \cellcolor[HTML]{FFDFDF}-0.30     & \cellcolor[HTML]{FFE7E7}-0.12       & \cellcolor[HTML]{FFDFDF}-0.30         & \cellcolor[HTML]{FFF8F8}0.26          \\
            & \textbf{CoT}                                                      & \cellcolor[HTML]{DAEDD3}0.99        & \cellcolor[HTML]{FAFDF9}0.49   & \cellcolor[HTML]{E4F2DF}0.83    & \cellcolor[HTML]{ECF6E8}0.71     & \cellcolor[HTML]{FFFFFE}0.42         & \cellcolor[HTML]{FFD0D0}-0.64        & \cellcolor[HTML]{FCFEFB}0.47          \\
        \multirow{-3}{*}{\begin{tabular}[c]{@{}l@{}}LLM \\  Reasoning\end{tabular}}            & \textbf{Reflexion}                                                  & \cellcolor[HTML]{BEDFB2}1.42        & \cellcolor[HTML]{FBFDFA}0.48 & \cellcolor[HTML]{E9F4E4}0.76      & \cellcolor[HTML]{D9ECD2}1.00        & \cellcolor[HTML]{FDFEFD}0.44         & \cellcolor[HTML]{FFD5D5}-0.53        & \cellcolor[HTML]{F3F9F1}0.60           \\ \midrule
            & \textbf{T=1}                                                    & \cellcolor[HTML]{6EB754}2.66        & \cellcolor[HTML]{EAF5E6}0.74   & \cellcolor[HTML]{63B247}2.83      & \cellcolor[HTML]{B2D9A4}1.61     & \cellcolor[HTML]{FFE8E8}-0.10        & \cellcolor[HTML]{BCDEB0}1.45         & \cellcolor[HTML]{B7DBAA}1.53 \\
            & \textbf{T=2}                                                    & \cellcolor[HTML]{8AC575}2.47        & \cellcolor[HTML]{EAF5E6}0.74   & \cellcolor[HTML]{4EA72E}3.39      & \cellcolor[HTML]{A1D190}1.61     & \cellcolor[HTML]{FFE6E6}-0.10       & \cellcolor[HTML]{BCDEB0}1.44          & \cellcolor[HTML]{C3E1B8}\textbf{1.59}          \\
        \multirow{-3}{*}{\begin{tabular}[c]{@{}l@{}}BIDDER\\  (\textbf{Ours})\end{tabular}} & \textbf{T=3}                                                    & \cellcolor[HTML]{B2D9A4}1.61        & \cellcolor[HTML]{EAF5E6}0.74  & \cellcolor[HTML]{68B44C}2.76     & \cellcolor[HTML]{A1D190}1.87     & \cellcolor[HTML]{FFE6E6}-0.14        & \cellcolor[HTML]{FFFEFE}0.39         & \cellcolor[HTML]{CCE6C3}1.21   \\ \bottomrule      
    \end{tabular}
    }
        \caption{The outcomes (payoff) of various models against different opponents in Limit Texas Hold'em. To more clearly distinguish between varying degrees of Payoff, the background color will be greener as the profit increases and redder as it decreases.}
    \label{tab:payoff}
\end{table*}

%% file: tabel/tab_nash.tex
\begin{table}[ht]
\centering
\resizebox{0.8 \textwidth}{!}{
    \begin{tabular}{@{}l|l|cccccc|c@{}}
    \toprule
   \% & P vs O           & \textit{Random} & \textit{Rule}  & \textit{CFR} & \textit{DQN} & \textit{DMC}  & \textit{DeepCFR} & AVG   \\ \midrule
    \multirow{3}{*}{Rule}                                                                    & \textbf{Random}  & 33.72           & 37.76      & 29.65       & 34.52        & 34.22            & 32.52            & 33.73          \\
    & \textbf{Rule}    & 35.29           & 36.44       & 44.81       & 48.36        & 43.54           & 42.08            & 41.75 
           \\
    & \textbf{CFR}     & 46.43           & 51.43 & 37.77         & 39.38        & 42.22               & 40.82            & 43.01
    \\ \midrule
    \multirow{2}{*}{\begin{tabular}[c]{@{}l@{}}Reinforcement \\ Learning\end{tabular}}       & \textbf{DQN}     & 54.86           & 54.59   & 60.45            & 51.66        & 64.48          & 52.65            & 56.45          \\
    & \textbf{DMC}     & 42.00              & 50.62      & 43.51       & 29.36        & 39.3             & 28.95            & 38.95             \\ \midrule
    \multirow{2}{*}{\begin{tabular}[c]{@{}l@{}}LLM \\ Reasoning\end{tabular}}                & \textbf{Direct}  & 6.20             & 2.01  & 6.90         & 5.56         & 7.85                  & 2.43             & 5.16           \\
    & \textbf{CoT}   & 9.32            & 2.05   & 11.43          & 11.6         & 15.6              & 6.96             & 9.49           \\ \midrule
    \multirow{3}{*}{\begin{tabular}[c]{@{}l@{}}BIDDER\\  (\textbf{Ours})\end{tabular}} & \textbf{T=1} & 65.51           & 55.75    & 66.76        & 47.29        & 80.81             & 28.68            & 57.47          \\
    & \textbf{T=2} & 68.38           & 58.62     & 70.09        & 45.77        & 79.24            & 29.79            & \textbf{58.97} \\
    & \textbf{T=3} & 66.03           & 55.75  & 69.85         & 45.59        & 78.65              & 28.98            & 57.47          \\ \midrule
    \textbf{Equilibrium}                                                                     & \textbf{DeepCFR} \tablefootnote{Due to the presence of random perturbations in the calculation of CFR, the results are not 100\% consistent.} & 91.99           & 91.07  & 93.7         & 89.45        & 93.15               & 83.69            & 90.51          \\ \bottomrule
    \end{tabular}
    }
    \vspace{2mm}
    \caption{
The Rational Degree (\%) of various models in Limit Texas Hold'em against different opponents, as calculated by Equation \ref{eq:rational}.}

    \vspace{-4mm}
    \end{table}

%% file: tabel/tab_neg.tex
\begin{table}[ht]
\resizebox{\textwidth}{!}{
\begin{tabular}{@{}l|ccccc|c@{}}
\toprule
\textbf{\begin{tabular}[c]{@{}l@{}}\diagbox{\textbf{Player}}{\hspace{-1cm}\textit{Opponent}}\end{tabular}}                      & \textit{MCTS} & \textit{Direct} & \textit{CoT} & \textit{ToT} & \textit{BIDDER} & Avg \\
\midrule

\textbf{MCTS}        & [14.8, 100\%, 60\%] & [15.9, 100\%, 20\%] & [22.4, 100\%, 30\%] & [18.5, 100\%, 17\%] & [15.0, 100\%, 33\%] & [17.3, \textbf{100\%}, \textbf{32\%}] \\
\textbf{Direct}       & [15.9, 100\%, 20\%] & [11.2, 50\%, 40\%]  & [16.2, 50\%, 40\%]  & [11.2, 50\%, 60\%]  & [21.0, 17\%, 0\%]   & [15.1, 53\%, \textbf{32\%}]  \\
\textbf{CoT}           & [25.2, 100\%, 30\%] & [22.0, 50\%, 40\%]  & [22.4, 70\%, 29\%]  & [16.4, 80\%, 50\%]  & [23.0, 67\%, 0\%]   & [21.8, 73\%, 30\%]  \\
\textbf{ToT}          & [20.8, 100\%, 17\%] & [15.0, 50\%, 60\%]  & [16.4, 80\%, 50\%]  & [19.8, 71\%, 20\%]  & [21.5, 67\%, 0\%]   & [18.7, 74\%, 29\%]  \\
\textbf{BIDDER(Ours)} & [30.3, 100\%, 33\%] & [21.0, 17\%, 0.0\%]   & [18.0, 67\%, 0\%]   & [17.0, 67\%, 0\%]   & [24.0, 50\%, 33\%]     & [\textbf{22.6}, 60\%, 13\%]   \\

\bottomrule
\end{tabular}
}
    \caption{Performance of various models in Negotiation against different opponents, and each cell represents [\textbf{Score, Agreement, Pareto Optimality}]}
    \vspace{-3mm}
    \label{tab:neg}
\end{table}